# AraEngage: Interpretable Linguistic Correlates of Engagement in Arabic Facebook News

**Shimaa Ibrahim[1], Nour Elhouda Dalloua[1], Md. Rafiul Biswas[2], Wajdi Zaghouani[1]**

[1]Northwestern University in Qatar, Doha, Qatar
[2]Hamad Bin Khalifa University, Doha, Qatar
{shimaa.ibrahim, wajdi.zaghouani}@northwestern.edu,
nourelhoudadalloua@gmail.com, mbiswas@hbku.edu.qa

## Abstract

Engagement features developed for English are routinely applied to Arabic, but unvocalized Arabic changes what a lexical feature measures. We introduce **AraEngage**, a querydefined sample of 10,000 public Arabic Facebook news posts from 2,451 pages in 57 countries (2023–2024) that pairs engagement counts with interpretable linguistic features. We use it to examine which interpretable Arabic indicators are associated with follower-normalized engagement, and how far those associations can be trusted given publisher context, the definition of engagement, and lexical ambiguity. Religious expressions, message length, sentence length, and lexical density are the leading correlates in a Random Forest model ($R^2$ = 0.38 on a post-level split) interpreted with SHAP. Unlike English news, positive sentiment is associated with higher engagement in this Arabic news sample. Message length shows a *length paradox*: unadjusted raw engagement is highest for long posts, whereas the model's attributions for follower-normalized engagement favor short posts. A linguistic audit identifies why standard lexicon features are fragile in unvocalized Arabic: imperative lists contain homographs such as عبر ("express!", but also the frequent preposition "via"), hedge lists conflate the possibility marker قد + imperfect ("may") with the emphatic قد + perfect ("has indeed"), and religious lists mix devotional formulas with place names in political news. These are plausible sources of measurement error that may attenuate the nearzero associations of imperatives and hedges. All findings are observational associations with operationalized indicators. We release the derived linguistic features, engagement counts, refined lexicons, and analysis code; raw posts are not redistributed, in line with CrowdTangle terms.

## 1 Introduction

Likes, shares, and comments are widely used as proxies for how content resonates with audiences (Berger and Milkman, 2012; Brady et al., 2017; Mahmud et al., 2014). A long line of work asks how *wording*, not only topic or author, relates to propagation (Tan et al., 2014; Ashok et al., 2013; Xu and Chawla, 2017), and finds associations with emotional valence, informativeness, readability, and platform conventions (Guerini et al., 2011; Kramer et al., 2014). Almost all of this work is on English. Arabic studies examine single features, such as hedging (Soufan, 2026), or single dialects (Ettalibi and El Ouardighi, 2025).

Transferring English feature sets to Arabic is harder than it looks, because unvocalized script makes many surface forms ambiguous. An "imperative" lexicon that lists عبر ("express!") also counts every occurrence of the preposition "via", one of the most frequent words in news text. A "hedge" lexicon that lists قد counts both قد يكون("may be") and قد قال ("has indeed said"), so it mixes uncertainty with its opposite. Such features can be computed without error and still measure something other than what their name says. Arabic also lacks the indefinite article that English work uses to measure "generality", and readability formulas such as Flesch are not defined for it (El-Haj and Rayson, 2016).

We therefore frame the study around the reliability of interpretable indicators, not only their predictive value. **RQ1**: Which textual and contextual indicators are associated with follower-normalized engagement on Arabic news posts? **RQ2**: Which English-derived indicators can be transferred to Arabic as valid measurements, and which are at risk of measuring something else? **RQ3**: Which conclusions depend on how engagement is defined and on publisher context?

**Contributions.** (1) **AraEngage**, a multi-country feature release for Arabic Facebook news posts containing engagement counts, page context, nine derived linguistic features per post, and a datasheet (§3, Appendix A); raw post text is not redistributed. (2) A **linguistic audit** identifying homographs, the two functions of قد, and topical place names as sources of measurement error in Arabic lexicon features, with a feature-transfer table, refined lexicons, and guidance for anyone transferring English feature sets (§5.4). (3) An analysis of the **length paradox**, in which descriptive and model-based evidence on message length point in opposite directions, and of what the existing evidence can and cannot separate (§5.3). (4) Evidence that the **negativity effect** reported for English news does not hold in this Arabic sample, together with the leading correlates of engagement: religious expressions, sentence length, and lexical density (§5.2).

## 2 Related Work

**Wording and engagement.** Propagation depends on wording as well as on topic and author (Tan et al., 2014). Informativeness, part-of-speech profiles, and explicit requests to share predict shareability (Ashok et al., 2013; Xu and Chawla, 2017). Emotional arousal of either valence is associated with sharing (Berger and Milkman, 2012; Godes et al., 2005; Heath et al., 2001; Hansen et al., 2011). Emotions spread through networks (Kramer et al., 2014), especially in moral and political discourse (Brady et al., 2017), and negativity attracts attention in news contexts (Guerini et al., 2011; Heiss et al., 2019). LIWC categories for cognitive and social processes also predict engagement (Mahmud et al., 2014).

**Arabic engagement and persuasion.** Soufan (2026) associate hedges (ر بما, قد) with engagement, and Ettalibi and El Ouardighi (2025) study Moroccan Arabic. Suchan (2014) describe strong emotional expression and metaphor as features of Arabic persuasion, and Myers et al. (2023) discuss religiosity in online engagement. These studies examine single features or single varieties; none combines them in one model or examines how Arabic lexical features behave under transfer.

**Orthographic ambiguity in Arabic.** Arabic is usually written without short vowels, so a single undiacritized form can correspond to several words with different parts of speech, and clitics attach conjunctions, prepositions, and pronouns to their hosts (Habash, 2010; Darwish et al., 2021). Arabic NLP addresses this with segmenters and in-context morphological disambiguators such as MADAMIRA,

| | |
|---|---|
| Posts | 10,000 public Facebook page posts |
| Pages | 2,451, administered from 57 countries |
| Period | 2023-04-12 to 2024-08-07 |
| Retrieval | CrowdTangle, five MSA news keywords |
| Engagement | reactions by type, comments, shares, followers at posting, Interactivity Index |
| Context | page category, post type, page-admin country |
| Linguistic features | length, lexical density, numerals, hashtags, sentence length, sentiment label, religious, imperative, and hedge counts |
| Resources | lexicons (original and refined), feature-extraction and analysis code |
| Privacy | no page names or user identifiers |

Table 1: AraEngage at a glance.

Farasa, and CAMeL Tools (Pasha et al., 2014; Abdelali et al., 2016; Obeid et al., 2020). Engagement studies, by contrast, typically count surface forms from word lists. Whether such counts measure their intended construct in Arabic has not been examined; our audit addresses this gap.

**Arabic language technology and interpretation.** Arabic sentiment classification relies on pre-trained encoders such as CAMeLBERT (Inoue et al., 2021), AraBERT (Antoun et al., 2020), and MARBERT (Abdul-Mageed et al., 2021). For interpretation, SHAP (Lundberg and Lee, 2017; Lundberg et al., 2020) attributes predictions to features. Attributions describe a fitted model and are complemented by partial dependence views (Friedman, 2001).

## 3 The AraEngage Dataset

We retrieved public Facebook page posts through CrowdTangle, Meta's research tool (discontinued in August 2024), with the Arabic keywords أخبار (news), أخبار عاجلة (breaking news), عناو ين الأخبار(news headlines), تحديثات (updates), and الأحداث الجار ية(current events), between 2023-04-12 and 2024-08-07. The export contains 10,000 posts from 2,451 pages whose administrators are located in 57 countries. Page-admin country is a platform field and does not identify audience location. The 41 export fields are listed in Table 2; we use the text fields (Message, Link Text, Image Text), page and post metadata, and engagement counts. Table 1 summarizes the dataset, and §7 describes its public release.

Generic keywords also retrieve non-news content. تحديثات (updates), for example, matches app and platform announcements, which is reflected in frequent tokens such as تطبيق (app), منصة(platform), and تيليجرام (Telegram) (Table 3). We keep these posts, since they are part of what a keyworddefined news corpus contains, and treat the corpus as a query-defined sample rather than a sample of news journalism.

emotional reaction types (all reactions except Like; Figure 4), Love is the most frequent (57.1%), followed by Haha (21.2%), Sad (11.3%), and Care (7.3%); Angry and Wow are rare. By page category (Figure 1, right), total raw engagement is highest for government organizations and movie writers, followed by general activities, services, personal blogs, and education pages, while science, app, car, newspaper, and entertainment pages register relatively little. Because these are totals, they reflect how many posts each category contributes as well as how much engagement each post receives.

Table 3 shows the most frequent keywords. Besides the query word اخبار (news), they fall into three groups. Religious: الله (God). Political: فلسطين (Palestine), عاجل (urgent/breaking). Platform and commercial: رابط (link), منصة (platform), تيلجرام

| Category | Description | Columns |
|---|---|---|
| Page-level (7) | Attributes of the source page | Page Name; User Name; Facebook Id; Page Category; Page Admin Top Country; Page Description; Page Created |
| Page audience at posting (2) | Page size when posted | Likes at Posting; Followers at Posting |
| Post timing (3) | Timestamps | Post Created; Post Created Date; Post Created Time |
| Post type (1) | Declared type | Type (Video, Status, Photo, Link) |
| Engagement totals (4) | Aggregate counts | Total Interactions; Likes; Comments; Shares |
| Emotional reactions (6) | Reaction counts | Love; Wow; Haha; Sad; Angry; Care |
| Video-related (6) | Video metadata | Video Share Status; Is Video Owner?; Post Views; Total Views; Total Views For All Crossposts; Video Length |
| Links & URLs (3) | References | URL; Link; Final Link |
| Content & text (4) | Text fields | Message; Image Text; Link Text; Description |
| Sponsorship (3) | Sponsor fields | Sponsor Id; Sponsor Name; Sponsor Category |
| Performance (2) | CrowdTangle indicators | Total Interactions (weighted); Overperforming Score |

Table 2: The 41 CrowdTangle export columns.

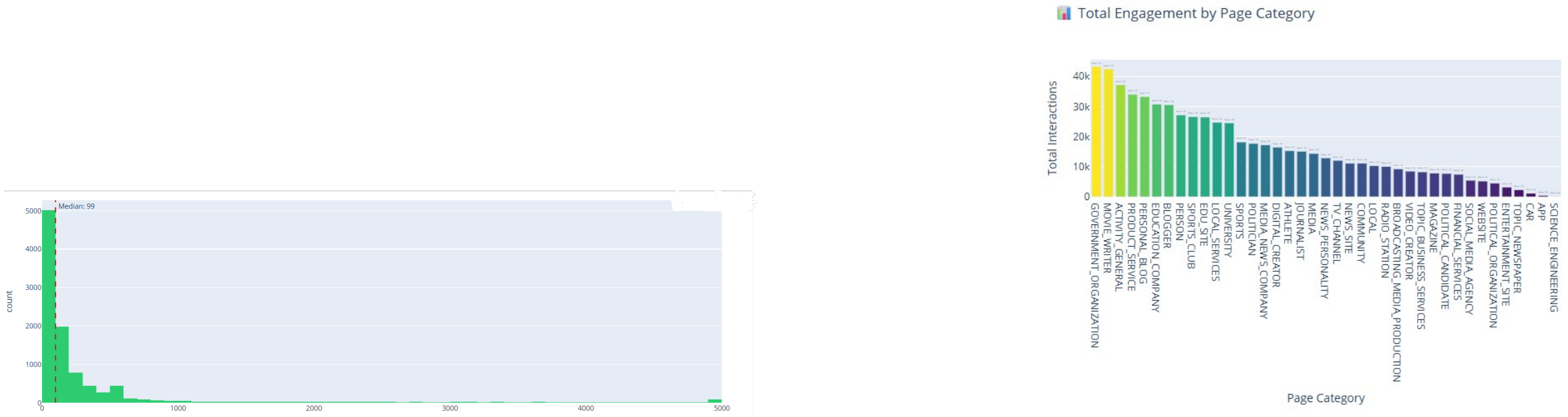

Figure 1: Left: distribution of message length (characters). Right: total raw engagement by page category; totals combine the number of posts per category with per-post engagement.

**Corpus profile.** Messages are short and rightskewed: the mean length is 349 characters, the median 99, and the maximum 4,999 (Figure 1, left). Only 16.3% of posts contain a URL, and posts average 0.75 hashtags, 0.20 exclamation marks, and 0.09 question marks, a register closer to assertive announcement than to conversation. Sharing is rare relative to liking (share-to-like ratio: mean 0.071, median 0), and engagement is concentrated, with the top decile of posts beginning at 180 interactions. Among the (Telegram), لمتابعة (to follow). The third group reflects the generic query term تحديثات (updates) and shows that a keyword-defined news corpus mixes journalism with channel promotion and app announcements. عبر (via) is the second most frequent keyword, which becomes important in §5.4. Among frequent keywords, الله has the highest average raw engagement (133.1 interactions, vs. 90.9 for اخبار), while political terms such as فلسطين (26.7) and platform terms such as منصة (26.4) receive the least.

| Arabic | English | Freq. | Avg. eng. | Total eng. |
|---|---|---|---|---|
| اخبارعبررابط | news via / through link | 4,739<br>845<br>838 | 90.94<br>51.79<br>72.35 | 430,954<br>43,764<br>60,633 |
| الله | God | 782 | 133.08 | 104,068 |
| متابعة | to follow | 731 | 69.62 | 50,891 |
| عاجل | urgent / breaking | 684 | 45.51 | 31,131 |
| فلسطين | Palestine | 593 | 26.72 | 15,844 |
| منصة | platform | 563 | 26.41 | 14,870 |
| تيلجرام | Telegram | 493 | 26.26 | 12,948 |
| اخر | other / last | 490 | 79.59 | 38,998 |

Table 3: Most frequent keywords in AraEngage with average and total raw interactions (not followernormalized). Unvocalized words can have several readings; the English column gives the main ones.

# 4 Method

## 4.1 Target: Interactivity Index

Following the weighting used by Sotrender and Vairetti et al. (2026), and the follower normalization of Galeazzi et al. (2024), we define

$$\text{InI} = \frac{1 \cdot \text{Reactions} + 5 \cdot \text{Comments} + 8 \cdot \text{Shares}}{\text{Followers at posting}},$$

where Reactions sums all reaction types including Like. Clicks are not available. Dividing by followers compares posts from pages of very different sizes. A Yeo-Johnson transform (Yeo and Johnson, 2000) reduces the strong right skew. The weights are a pragmatic convention that ranks comments and shares as costlier forms of engagement than reactions; they are not a validated behavioral scale (§5.6).

## 4.2 Features

Table 4 summarizes the features. All text fields are concatenated. **Message length** is the number of characters. **Lexical density** is the share of contentword tokens (nouns, proper nouns, verbs, adjectives, adverbs, numerals) among all tokens, tagged with Stanza's tokenize,mwt,pos pipeline (Qi et al., 2020). It approximates information load (Halliday, 1985); proper nouns are included because named actors are central to political news in the region (Baker, 2006). **Average sentence length** (tokens per sentence) is a coarse processing-load proxy, not a readability score: Flesch is not defined for Arabic. **Numerals** count Western and Arabic-Indic digit sequences, and **hashtags** count # tokens. **Sentiment** is the label (positive, negative, neutral) of the CAMeLBERT sentiment classifier (Inoue et al., 2021; Obeid et al., 2020), applied to the first 512 characters of each post. This is a character limit, stricter than the model's 512-subword input limit; because the median post has 99 characters, it affects only the longer posts, for which sentiment reflects the opening of the text. **Imperatives**, **religious expressions**, and **hedges** are counted with the lexicons in Table 6. Because Arabic religious formulas such as إن شاء الله pervade everyday language, we treat the religious count as a broad religion-related lexical signal, not a measure of an author's religiosity. The contextual features are page category, post type, and followers at posting.

| **Group** | **Operationalization** | **Motivation** |
|---|---|---|
| Length | number of characters | Tan et al. (2014) |
| Lexical | content-word density (Stanza POS); numeral count | Ashok et al. (2013); Xu and Chawla (2017) |
| Processing load | mean tokens per sentence | Tan et al. (2014) |
| Sentiment | CAMeLBERT label (pos/neg/neu) | Berger and Milkman (2012) |
| Platform | hashtag count | Xu and Chawla (2017) |
| Religious | religiousexpression lexicon count | Myers et al. (2023); Suchan (2014) |
| Call to action | imperative lexicon count | Tan et al. (2014) |
| Hedging | epistemichedge lexicon count | Soufan (2026) |
| Context | page category, post type, followers | (contextual controls) |

Table 4: Features. “Generality”, used in English work through the indefinite article, is not operationalized: Arabic has no indefinite article, and definiteness alone is not a valid proxy.

| Model | MSE | $R^2$ |
|---|---|---|
| Ridge | 0.8035 | 0.2337 |
| MLP | 0.7456 | 0.2888 |
| $k$-NN | 0.7028 | 0.3297 |
| XGBoost | 0.6728 | 0.3583 |
| Random Forest | **0.6473** | **0.3826** |

Table 5: Prediction of the transformed, followernormalized InI on the held-out 20% split.

### 4.3 Models and Interpretation

We compare Ridge regression, $k$-nearest neighbors, Random Forest (Breiman, 2001), XGBoost (Chen and Guestrin, 2016), and a one-hidden-layer MLP (Hastie et al., 2009), implemented in scikit-learn (Pedregosa et al., 2011). Data are split 80/20 at random (seed 42). Numeric features are standardized on the training set, and categorical features are one-hot encoded. SHAP values (Lundberg and Lee, 2017) are computed for the best model. **How to read the numbers.** $R^2$ and mean squared error (MSE) are computed on the *transformed, followernormalized* InI, not on raw counts. The split is at the post level, so posts from the same page can appear in both training and test data. SHAP values describe the fitted model’s use of each feature; they indicate associations, not causal effects.

## 5 Results

### 5.1 Model performance

Tree ensembles perform best (Table 5): Random Forest explains 38% of the variance of the transformed InI, XGBoost 36%, and Ridge regression

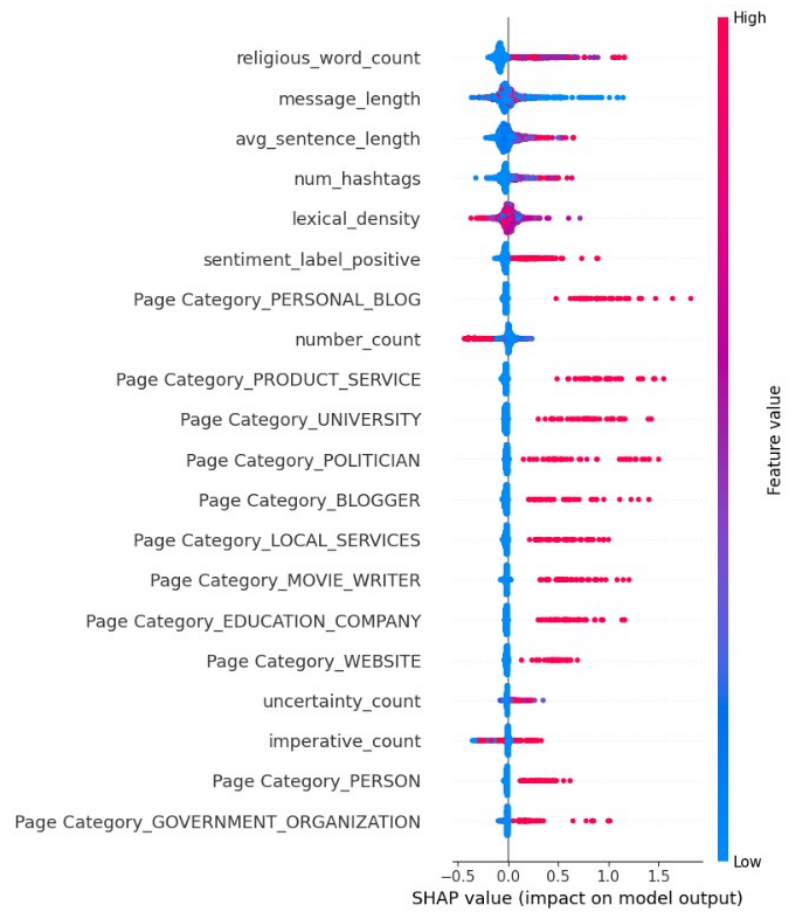

Figure 2: SHAP values for the Random Forest model (test set). Each point is a post; color shows the feature value. Page categories with positive contributions include personal blogs, product/service, university, politician, and blogger pages; government-organization pages have lower contributions.

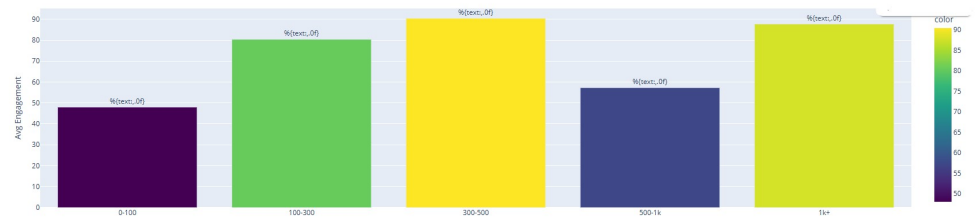

Figure 3: Average raw engagement by message-length bin (characters), without conditioning on page or other features.

23%. The gap to Ridge indicates that part of the association is non-linear or involves interactions. The rest of the analysis interprets the Random Forest model.

### 5.2 Leading correlates

The four most influential features are textual (Figure 2). Higher **religious-expression** counts have larger positive SHAP values. In the raw data, الله(God) has the highest average engagement among frequent keywords (133.1 interactions per occurrence vs. 90.9 for اخبار, news; Table 3). Higher **sentence length** and **lexical density** also have positive contributions: posts that pack more content into longer sentences receive higher predicted engagement. **Message length** is the second most influential feature, with the direction discussed in §5.3.

### 5.3 The length paradox

Average raw engagement is highest for posts of 300–500 and over 1,000 characters (Figure 3), whereas the model’s SHAP attributions favor *shorter* posts. Two things differ between these analyses at the same time. **The engagement measure.** The binned averages use raw interaction counts; the model uses follower-normalized, transformed lnI.

**The adjustment.** The binned averages are unadjusted, whereas SHAP attributes the model's prediction across all features jointly, including page category, post type, and followers. Either difference can produce the reversal. For example, if larger pages publish longer posts, long posts would have high raw counts because of their audiences but not high engagement relative to followers. The existing evidence cannot isolate which of the two differences is responsible. Nor does a SHAP attribution establish that shortening a post would change its engagement: the result holds within the fitted model's attribution framework. In short: *unadjusted raw engagement favors long posts; the model's attributions for follower-normalized engagement favor short posts. The two measure different things.* Any claim about message length should therefore state its engagement measure and its adjustment set; separating the two requires unadjusted follower-normalized engagement by length bin and an adjusted effect estimate such as accumulated local effects.

### 5.4 Linguistic audit: what may not transfer

Imperative and hedge counts have SHAP values close to zero, and the religious count is the most influential feature. We audited all three lexicons (Table 6) for sources of measurement error that bear on these results. Table 7 lists the error sources with constructed examples.

**Imperatives are dominated by homographs.** عبر ("express!") is also the preposition "via / through". It occurs 845 times in the corpus and is among the most frequent keywords (Table 3). طالب("demand!") is also "student", دافع ("defend!") "motive", صوت ("vote!") "voice", and تابع("follow!") "affiliated with". شارك, علق, and قاطعalso read as third-person past-tense verbs ("he shared", "he commented", "he boycotted"). In news text the nonimperative readings are likely to dominate, so the "imperative" count mostly measures prepositions and reported actions. **Hedges conflate uncertainty with emphasis.** With an imperfect verb, قد marks possibility ( قد يكون, "may be"); with a perfect verb, it marks completion or emphasis (قد قال, "has indeed said") (Ryding, 2005). News reporting uses the second construction often, so the hedge count mixes uncertainty with assertion. يمكن adds ability readings ("can") to epistemic ones.

| List | Lexicon used in this study | Refinement proposed by the audit (§5.4) |
|---|---|---|
| Imperative | شارك، انشر، أعد، عمم، ارسل، علق، اكتب، قل، عبر، صوت، شاهد، تابع، اقرأ، اطلع، لاحظ، تحرك، سارع، انضم، ادعم، قاطع، قاوم، طالب، ارفض، دافع، ندد<br>*Share, publish, repeat, spread, send, comment, write, say, express, vote, watch, follow, read, look, notice, move, hurry, join, support, boycott, resist, demand, reject, defend, condemn* | Remove homographs عبر (via), طالب (student), دافع (motive), صوت (voice) أعد (prepared), تابع (affiliated), and past-tense homographs شارك ، علق، قاطع; add unambiguous plural forms شاركوا، تابعوا، تابعونا، علقوا، قاطعوا، صوتوا. |
| Religious | الله، اللهم، إن شاء الله، الحمد لله، استغفر الله، الإسلام، مسلم، النبي، الرسول، القرآن، دين، إيمان، دعاء، عبادة، تقوى، القدس، الأقصى، المسجد الأقصى<br>*Allah, O Allah, God willing, praise be to God, I seek God's forgiveness, Islam, Muslim, the Prophet, the Messenger, Quran, religion, faith, supplication, worship, piety, Jerusalem, Al-Aqsa, Al-Aqsa Mosque* | Count القدس، الأقصى، المسجد الأقصى separately as sacred places; replace دين (also "debt") with الدين; match longer formulas first so that الله is not counted twice. |
| Uncertainty | قد ، ربما، يمكن، من الممكن، من المحتمل *may/has, perhaps, can, it is possible, it is likely* | Count قد only before an imperfect verb; drop يمكن (ability); add يحتمل ، على الأرجح، لعل. |

,

Table 6: Lexicons used for feature extraction (middle) and refinements derived from the linguistic audit (right). Matching is applied after normalization of alef forms, alef maqsura, ta marbuta, diacritics, and tatweel.

| Feature | Form | Intended reading (example) | Competing reading (example) |
|---|---|---|---|
| Imperative | عبر | "express!": عبّر عن رأيك (express your opinion) | preposition "via": أعلن الوزير عبر حسابه الرسمي (the minister announced via his official account) |
| | طالب | "demand!": طالب بحقك (demand your right) | noun "student": طالب جامعي (a university student); past tense "demanded" |
| | تابع | "follow!": تابع آخر الأخبار (follow the latest news) | adjective "affiliated with": تابع لوزارة الصحة (affiliated with the Ministry of Health) |
| | شارك | "share!": شارك المنشور (share the post) | past tense "took part": شارك الرئيس في القمة (the president took part in the summit) |
| | دافع | "defend!": دافع عن حقوقك (defend your rights) | noun "motive": دافع الجريمة (the motive for the crime) |
| Hedge | قد | + imperfect, "may": قد يرتفع سعر النفط (oil prices may rise) | + perfect, "has (indeed)": قد أعلنت الوزارة (the ministry has announced) |
| | يمكن | epistemic "may": يمكن أن يكون السبب (the cause may be) | ability "can": يمكن للمواطنين التسجيل (citizens can register) |
| Religious | دين | "religion": الدين الإسلامي (the Islamic religion) | "debt": سداد الدين العام (repayment of public debt) |
| | القدس | sacred place in devotional text | place name in political news: مواجهات في القدس (clashes in Jerusalem) |
| | الله | devotional reference | counted again inside formulas such as إن شاء الله (God willing) |

Table 7: Sources of error in the AraEngage lexicons, with constructed illustrative examples. In unvocalized text, both readings have the same surface form and are counted by the lexicon.

**Religious counts mix devotion and topic.** The religious lexicon combines several phenomena: devotional formulas and invocations (إن شاء الله, الحمد لله, اللهم), general religious vocabulary (القرآن, الإسلام, إيمان), sacred-place names that frequently occur in political news (القدس, الأقصى), and forms that need context to disambiguate (دين, also "debt"). الله is listed both alone and inside formulas, so a single formula can contribute more than once. The religious association in §5.2 therefore concerns this composite indicator. It cannot be attributed to devotional language alone, and the relative contribution of the components is not identified by the current feature.

**Implication for feature transfer.** Non-target matches add noise to a count, and noise in a predictor tends to attenuate its estimated association. These ambiguities are therefore plausible sources of measurement error that may contribute to the nearzero attributions of imperatives and hedges. Their actual contribution remains unquantified until the refined features are evaluated. What the audit does establish is that the imperative and hedge counts are *not valid measures of their intended constructs as specified*, so their null results should not be read as evidence that calls to action or hedging are unrelated to engagement. Anyone transferring an English feature set to Arabic should (i) check each lexicon entry for unvocalized homographs, (ii) encode the syntactic context that disambiguates function words such as قد, (iii) separate topical from stylistic vocabulary, and (iv) report the most frequent matched forms. The refined lexicons in Table 6 implement (i)–(iii).

**What transfers, and what does not.** Table 8 summarizes how each English engagement feature fared. Four kinds of outcome emerge. Some features *transfer* once the right Arabic tool is used (lexical density with an Arabic POS tagger, sentiment with an Arabic classifier). Some must be *adapted*: word counts depend on how clitics are segmented, so we count characters; Flesch is undefined, so we use sentence length as a proxy; numerals must cover two digit systems. One feature *does not transfer* at all: generality has no Arabic counterpart to the indefinite article. And lexicon features for calls to action and hedging are *at risk*: they can be computed, but their surface forms also match other words and constructions. This outcome is the most dangerous, because nothing in the feature values signals the problem.

| English feature | Arabic issue → AraEngage choice | Outcome |
|---|---|---|
| Length (words) | clitics merge words → characters | adapted |
| POS profile | morphology → Stanza Arabic tagger | transfers |
| Readability | Flesch undefined → sentence length | adapted (proxy) |
| Generality | no indefinite article → omitted | does not transfer |
| Numerals | two digit systems → both counted | adapted |
| Hashtags | none | transfers |
| Sentiment | Arabic classifier needed → CAMeLBERT | transfers (not validated here) |
| Call to action | imperative homographs → lexicon | at risk |
| Hedging | قد polysemy → lexicon | at risk |
| Religiosity | formulas and place names → lexicon | partial |

Table 8: Transfer of engagement features from English to Arabic news posts.

**5.5 Sentiment: no negativity effect** Positive sentiment is associated with higher predicted engagement. *Unlike English news, where negativity attracts attention (Heiss et al., 2019), positive sentiment is associated with higher engagement in this Arabic keyword-defined news sample.* One possible reason is the share of devotional, congratulatory, and condolence content in a keyworddefined news corpus, which also fits the religious association above. The reaction mix points the same way: Love is the most frequent emotional reaction (57.1% of non-Like reactions), ahead of Haha (21.2%), Sad (11.3%), and Care (7.3%) (Figure 4). Our data cannot separate these explanations.

**5.6 What these results do not show Causality.** The features are correlates, not causes; a SHAP value describes the fitted model's use of a feature, not the effect of editing a post. **Publisher transfer.** Because the split is at the post level, $R^2$ describes new posts from pages represented in training, not previously unseen pages, and the feature ranking comes from a single fitted model whose stability across splits is not quantified. **Follower coupling.** Followers at posting appear both in the denominator of InI and among the predictors, which can create a mechanical association between follower count and the target; the contribution of follower count should be read with this in mind. Followers are also not impressions or observed reach.
**Engagement definition.** The 1/5/8 weights are a substantive measurement choice; other weightings may change the ranking of features. **Constructs.** The religious count is a composite of devotional, lexical, and place-name matches, not a measure of the religiosity of authors or readers, and reactions are aggregate audience responses, not individual emotions. **Scope.** Results are limited to a keyworddefined CrowdTangle sample of public pages from 2023–2024.

## 6 Discussion

**Publisher context, measurement, and linguistic generalization.** Interpreting linguistic engagement predictors requires separating three sources of variation. *Publisher variation*: a post-level evaluation estimates performance where publishers appear in both training and test data; page category and follower count only partially characterize editorial style, audience composition, and posting practices, so the importance of a linguistic feature should not be read as independent of publisher context. *Engagement measurement*: follower normalization makes pages of different sizes comparable, but followers are not impressions, and weighting reactions, comments, and shares is a substantive decision rather than a neutral transformation; our findings concern the specified Interactivity Index. *Linguistic measurement*: the religious indicator shows the gap between lexical detection and discourse function, since Arabic religious vocabulary appears in devotional formulas, conventional expressions, reported speech, and geopolitical references. Model explanations are therefore associations with operationalized indicators, not evidence about communicative intentions or the effects of rhetorical strategies. Together, these distinctions argue for evaluating interpretable features on their stability, incremental contribution, and linguistic validity, not only on predictive accuracy, particularly when feature inventories are transferred from English to Arabic.

**Measurement before interpretation.** The most consequential results in this paper concern measurement, not engagement. Two of nine features are at risk of measuring something other than their intended construct, and a third combines several phenomena. A study that reported the near-zero SHAP values for imperatives and hedges as substantive findings would conclude that calls to action and hedging do not matter for Arabic news engagement, a conclusion our data cannot support. The same risk applies to any lexicon-based feature transferred from English to a language with unvocalized script or rich cliticization. For Arabic, the remedies are available: in-context morphological disambiguation (Pasha et al., 2014; Obeid et al., 2020) would separate imperative from past-tense and nominal readings, and a syntactic rule on the verb following قد would separate possibility from emphasis.

**Why positivity?** The positive-sentiment association, the strength of religious expressions, and the dominance of Love among reactions point in the same direction. A keyword-defined Arabic news corpus contains not only reports of events but also congratulations, condolences, prayers, and religious formulas attached to news, which may attract affiliative responses. The negativity effect documented for English (Heiss et al., 2019) may therefore depend on the genre mix of what counts as “news” on a platform as much as on language or culture. Disentangling these explanations requires topic and genre labels, which AraEngage’s released features make it possible to add. **Uses of AraEngage.** AraEngage supports engagement modeling with richer features, validation of automatic features against human judgments, and comparisons across countries and page types.

## 7 Dataset Release

AraEngage will be publicly released upon publication. The release includes the post-level feature table (engagement counts, followers, InI, page and post context, and the nine derived features of §4.2), the original and refined lexicons (Table 6), the feature-extraction and analysis code, and the datasheet (Appendix A). It abides by Meta's terms of service and data-sharing policies. Data are licensed under CC BY-NC 4.0 and code and lexicons under MIT; users must not attempt re-identification or use the data for surveillance, profiling, or commercial targeting.

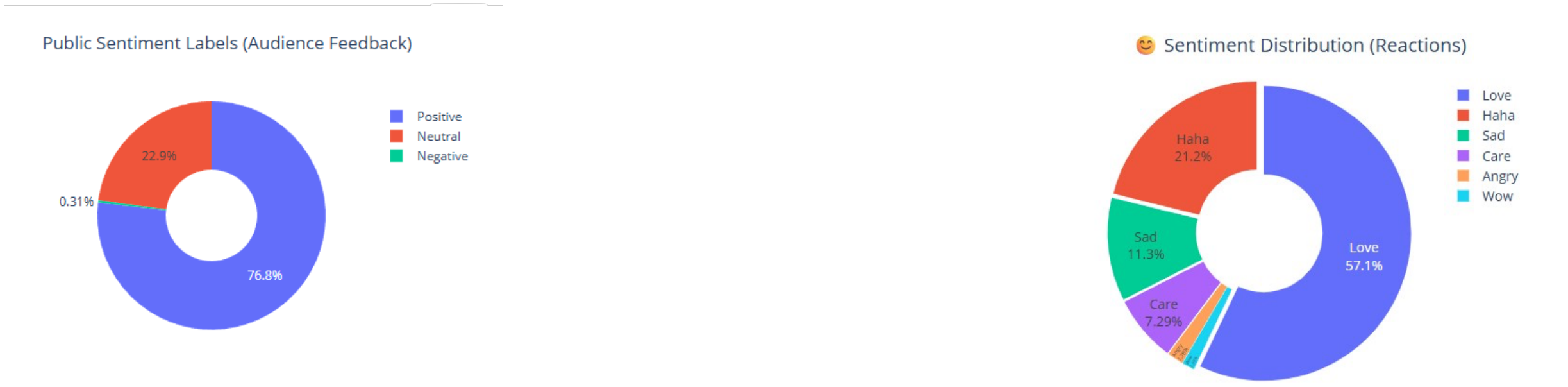

Figure 4: Left: CAMeLBERT sentiment labels of AraEngage posts. Right: distribution of emotional reactions (all reaction types except Like): Love 57.1%, Haha 21.2%, Sad 11.3%, Care 7.3%, with Angry and Wow below 2% each.

## 8 Conclusion

AraEngage pairs 10,000 Arabic Facebook news posts from 57 countries with engagement counts and interpretable features. With it, we find that religious expressions, sentence length, and lexical density are the leading correlates of followernormalized engagement, and that, unlike English news, positive sentiment is associated with higher engagement. Unadjusted raw engagement favors long posts, while the model's attributions for follower-normalized engagement favor short posts; the two analyses differ in both the engagement measure and the adjustment, and any claim about length should state both. **What transfers from English feature sets and what does not:** length, lexical density, and sentence length can be computed in Arabic without lexical ambiguity, and sentiment transfers given an Arabic classifier, whereas lexicon features for imperatives and hedges are at risk in unvocalized Arabic because of homographs and the two functions of قد, and the religious lexicon mixes devotion with topic. The central lesson is that interpretable Arabic indicators should be judged by whether they remain reliable once publisher context, the engagement definition, and lexical ambiguity are taken into account. Future work should evaluate generalization to unseen pages, test alternative engagement weights, separate the components of the religious indicator, validate the refined lexicons against human labels, and add Arabic readability measures (El-Haj and Rayson, 2016), modeling of code-switching and Arabizi (Nordenson, 2017), rhetorical devices such as metaphor (Alsiyat and Piao, 2020), and richer emotion categories (Kahn et al., 2007).

## Limitations

**Sampling.** The corpus is a 10,000-post CrowdTangle export retrieved with five MSA keywords. It over-represents formal news styles, includes nonnews posts matched by generic keywords, and under-represents dialectal, Arabizi, code-switched, and French- or English-language MENA news. **Measurement.** The lexicon features have the error sources described in §5.4, and the refined lexicons have not yet been validated against human labels. Sentiment is a three-way label from one classifier, applied without human validation on this corpus, and it cannot express mixed tone. Page category is a noisy self-declared field. **Evaluation.** Results come from a single random post-level split, so page-level style can be shared between training and test data, and the stability of the SHAP ranking across splits is not quantified. The InI weights are a convention whose alternatives were not tested. **Design.** All results are observational. Page-specific effects, posting time, and media content may confound the associations. **Scope.** Facebook pages only; 2023–2024; no comment text and no images or video.

## Ethical Considerations

**Data and access.** We analyze public page content retrieved through Meta's CrowdTangle research access. No private profiles, groups, or messages were accessed. We did not interact with users and make no individual-level inferences. The study involves computational analysis of publicly posted content and was classified as exempt from full IRB review under our institutional guidelines.

**Release.** AraEngage is released as described in §7, in accordance with Meta's terms of service and data-sharing policies. The release contains no page names or user identifiers, and examples in the paper are paraphrased.

**Risks.** Engagement correlates could be used to optimize content manipulatively, which we discourage. The religious feature must not be used to profile individuals or to infer religious commitment or extremism. The corpus covers politically sensitive topics, including Palestine, regional conflicts, and religious discourse; the model must not be used to rank or flag users or pages.

**Use of AI assistants.** AI assistants were used for writing refinement and literature search. All analyses and results were produced by the authors with the code described in §4.2.

## A Datasheet for AraEngage

Following Gebru et al. (2021), we summarize the dataset. **Motivation.** AraEngage was created to study which linguistic properties of Arabic news posts are associated with engagement, and to test how English engagement features transfer to Arabic. **Composition.** 10,000 public Facebook page posts from 2,451 pages whose administrators are located in 57 countries, published between 2023-0412 and 2024-08-07. Each instance is one post with engagement counts (reactions by type, comments, shares), followers at posting, page category, post type, page-admin country, and nine derived linguistic features (Table 4). The dataset contains no page names or user identifiers. **Collection.** Posts were retrieved through CrowdTangle with five MSA news keywords (§3). The collection is a keyword-defined sample and includes non-news posts matched by generic terms. **Preprocessing.** Text fields were concatenated before feature extraction; features were computed as described in §4.2. **Uses.** Engagementmodeling, featurevalidation, andcomparisons across countries and page types. AraEngage should not be used to profile individuals or pages, to infer religious commitment, or to optimize content for manipulative engagement. **Distribution.** Derived features under CC BY-NC 4.0; code and lexicons under the MIT License. **Maintenance.** The dataset is maintained by the authors; errata and updated lexicons will be published in the project repository.